\documentclass[10pts]{article}
\usepackage{graphicx}
\usepackage[round]{natbib}
\usepackage[margin=1.25in]{geometry}
\usepackage{todonotes}
\usepackage{url}

\begin{document}

\title{\textbf{Explanation Methods in Deep Learning:\\ Users, Values, Concerns and Challenges\footnote{This article will appear as a chapter in \textit{Explainable and Interpretable Models in Computer Vision and Machine Learning}, a Springer series on Challenges in Machine Learning.}}}

\author{Gabri\"{e}lle Ras, Marcel van Gerven, Pim Haselager \\ \\ Radboud University, Donders Institute for Brain, Cognition and Behaviour, \\ Nijmegen, the Netherlands \\ \texttt{\{g.ras, m.vangerven, w.haselager\}@donders.ru.nl}}
\date{}

\maketitle

\abstract{Issues regarding explainable AI involve four components: users, laws \& regulations, explanations and algorithms. Together these components provide a context in which explanation methods can be evaluated regarding their adequacy. The goal of this chapter is to bridge the gap between expert users and lay users. Different kinds of users are identified and their concerns revealed, relevant statements from the General Data Protection Regulation are analyzed in the context of Deep Neural Networks (DNNs), a taxonomy for the classification of existing explanation methods is introduced, and finally, the various classes of explanation methods are analyzed to verify if user concerns are justified. Overall, it is clear that (visual) explanations can be given about various aspects of the influence of the input on the output. However, it is noted that explanation methods or interfaces for lay users are missing and we speculate which criteria these methods / interfaces should satisfy. Finally it is noted that two important concerns are difficult to address with explanation methods: the concern about bias in datasets that leads to biased DNNs, as well as the suspicion about unfair outcomes.}

\section{Introduction}
Increasingly, Artificial Intelligence (AI) is used in order to derive actionable outcomes from data (e.g. categorizations, predictions, decisions). The overall goal of this chapter is to bridge the gap between expert users and lay users, highlighting the explanation needs of both sides and analyzing the current state of explainability. We do this by taking a more detailed look at each component mentioned above and in Figure \ref{fig:xai_components}. Finally we address some concerns in the context of DNNs.

\subsection{The components of explainability}
Issues regarding explainable AI (XAI) involve (at least) four components: users, laws and regulations, explanations and algorithms. Together these components provide a context in which explanation methods can be evaluated regarding their adequacy. We briefly discuss these components in Figure \ref{fig:xai_components}.

\subsection{Users and laws}
AI has a serious impact on society, due to the large scale adoption of digital automation techniques that involve information processing and prediction. Deep Neural Networks (DNNs) belong to this set of automation techniques and are used increasingly because of their capability to extract meaningful patterns from raw input. DNNs are fed large quantities of digital information that are easily collected from users. Currently there is much debate regarding the safety of and trust in data processes in general, leading to investigations regarding the explainability of AI-supported decision making. The level of concern about these topics is reflected by official regulations such as the General Data Protection Regulation (GDPR), also mentioned in \citep{doshi2017towards, holzinger2017we}, incentives to promote the field of explainability~\citep{gunning2017explainable} and institutional initiatives to ensure the safe development of AI such as OpenAI. As the technology becomes more widespread, DNNs in particular, the dependency on said technology increases and ensuring trust in DNN technology becomes a necessity. Current DNNs are achieving unparalleled performance in areas of Computer Vision (CV) and Natural Language Processing (NLP). They are also being used in real-world applications in e.g. medical imaging~\citep{lee2017fully}, autonomous driving~\citep{bojarski2017explaining} and legislation~\citep{lockett2017predictions}.

\subsection{Explanation and DNNs}
The challenge with DNNs in particular lies in providing insight into the processes leading to their outcomes, and thereby helping to clarify under which circumstances they can be trusted to perform as intended and when they cannot. Unlike other methods in Machine Learning (ML), such as decision trees or Bayesian networks, an explanation for a certain decision made by a DNN cannot be retrieved by simply scrutinizing the inference process. The learned internal representations and the flow of information through the network are hard to analyze: 
As architectures get deeper, the number of learnable parameters increases. It is not uncommon to have networks with millions of parameters. Furthermore, network architecture is determined by various components (unit type, activation function, connectivity pattern, gating mechanisms) and the result of a complex learning procedure, which itself depends on various properties (regularization, adaptive mechanisms, employed cost function). The net result of the interaction between these components cannot be predicted in advance.
Because of these complications, DNNs are often called \textit{black box models}, as opposed to glass-box models~\citep{holzinger2017glass}. Fortunately, these problems have not escaped the attention of the ML/Deep Learning (DL) community~\citep{zeng2016towards, samek2017explainable, seifert2017visualizations, olah2017feature, hall2017ideas, montavon2017methods, marcus2018deep, doshi2017towards}. 
Research on how to interpret and explain the decision process of Artificial Neural Networks (ANNs) has been going on since the late 1980's~\citep{elman1989representation, andrews1995survey}. The objective of explanation methods is to make specific aspects of a DNN's internal representations and information flow interpretable by humans. 
\label{sec:components}
\begin{figure}
    \centering
    \includegraphics[width=0.75\linewidth]{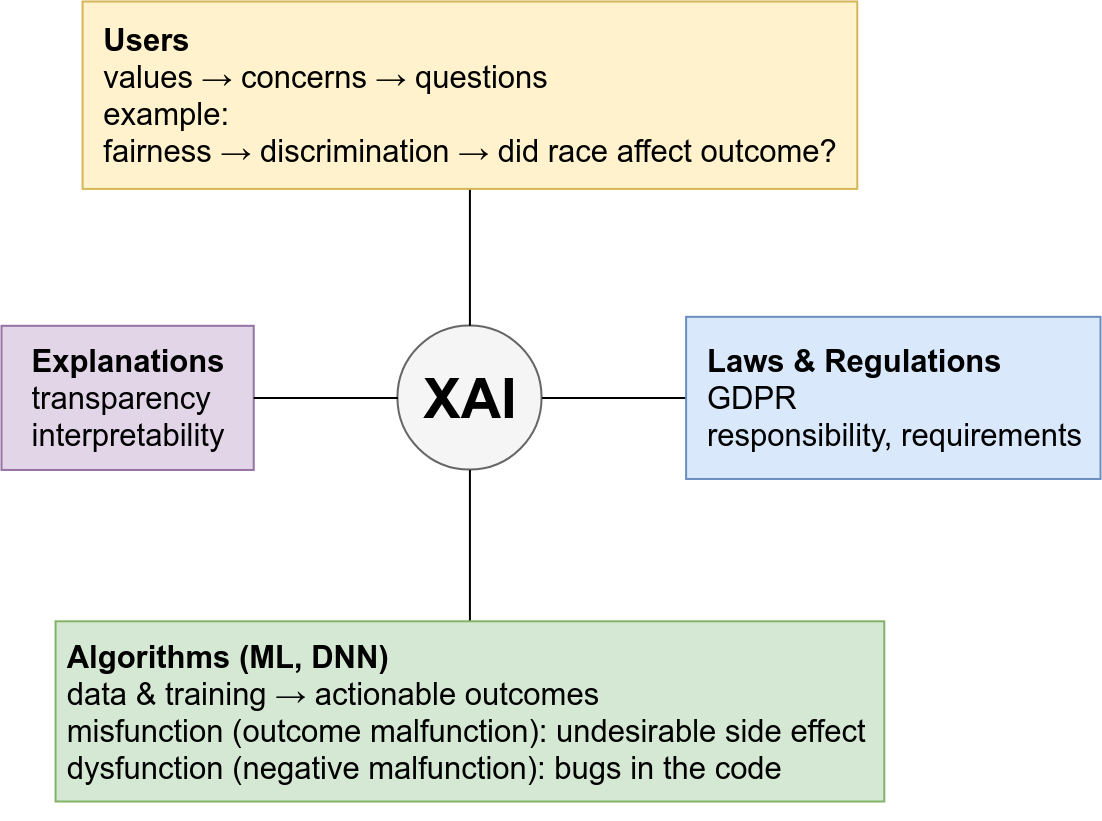}
    \caption{Issues regarding explainable DNNs involve (at least) four components: users, algorithms, laws and explanations. Together these components provide a context in which explanations can be evaluated regarding their adequacy.}
    \label{fig:xai_components}
\end{figure}

\section{Users and their concerns} 
\label{sec:users}
Various kinds of DNN users can be distinguished. Users entertain certain values; these include ethical values such as fairness, neutrality, lawfulness, autonomy, privacy or safety, or functional values such as accuracy, usability, speed or predictability. Out of these values certain concerns regarding DNNs may arise, e.g. apprehensions about discrimination or accuracy. These concerns get translated into questions about the system, e.g. ``did the factor ‘race’ influence the outcome of the system" or ``how reliable was the data used?" In this section we identify at least two general types of users: the expert users and the lay users, that can be further categorized into six specific kinds of users. Note that there could be (and there regularly is) overlap between the users described below, such that a particular user can be classified as belonging to more than one of the categories.
\begin{enumerate}
    \item \textbf{Expert users} are the system builders and/or modifiers that have direct influence on the implementation of the network. Two kinds of experts can be identified:
    \begin{enumerate}
        \item \textbf{DNN engineers} are generally researchers involved in extending the field and have detailed knowledge about the mathematical theories and principles of DNNs. DNN engineers are interested in explanations of a functional nature, e.g. the effects of various hyperparameters on the performance of the network or methods that can be used for model debugging.
        \item \textbf{DNN developers} are generally application builders who make software solutions that can be used by lay people. DNN developers often make use of off-the-shelf DNNs, often re-training the DNN along with tuning certain hyperparameters and integrating them with various software components, resulting in a functional application. The DNN developer is concerned with the goals of the overall application and assesses whether they have been met by the DNN solution. DNN developers are interested in explanation methods that allow them to understand the behavior of the DNN in the various use cases of the integrated software application. 
    \end{enumerate}
    \item \textbf{Lay users} do not and need not have knowledge of how the DNN was implemented and the underlying mathematical principles, nor do they require knowledge of how the DNN was integrated with other software components resulting in a final functional application. At least four lay users are identified:
    \begin{enumerate}
        \item \textbf{The owner} of the software application in which the DNN is embedded. The owner is usually an entity that acquires the application for possible commercial, practical or personal use. For example, an owner can be an organization (e.g. a hospital or a car manufacturer) that purchases the application for end users (e.g. employees (doctors) or clients (car buyers)), but the owner can also be a consumer that purchases the application for personal use. In the latter case the categorization of owner fully overlaps with the next category of users which are the end users. The owner is concerned with explainability questions about the capabilities of the application, e.g. justification of a prediction or a prediction given the input data, and aspects of accountability, e.g. to what extent can application malfunction be attributed to the DNN component?
        \item \textbf{The end user} for whom the application was intended to be used by. The end user uses the application as part of their profession or for personal use. The end user is concerned with explainability about the capabilities of the application, e.g. justification of a prediction given the input data, and explainability regarding the behavior of the application, e.g. why does the application not do what it was advertised to do?
        \item \textbf{The data subject} is the entity whose information is being processed by the application or the entity which is directly affected by the application outcome. An outcome is the output of the application in the context of the use case. Sometimes the data subject is the same entity as the end user, for example in the case that the application is meant for personal use. The data subject is mostly concerned with the ethical and moral aspects that result from the actionable outcomes. An actionable outcome is an outcome that has consequences or an outcome on which important decisions are based.
        \item \textbf{Stakeholders} are people or organizations without a direct connection to either the development, use or outcome of the application and who can reasonably claim an interest in the process, for instance when its use runs counter to particular values they protect. Governmental and non-governmental organizations may put forward legitimate information requests regarding the operations and consequences of DNNs. Stakeholders are often interested in the ethical and legal concerns raised in any phase of the process. 
    \end{enumerate}
\end{enumerate}

\subsection*{Case study: autonomous driving}
In this section the different users are presented in the context of a self-driving car. 
\begin{enumerate}
    \item The DNN engineer creates a DL solution to the problem of object segmentation and object classification by experimenting with various types of networks. Given raw video input the DL solution gives the output of the type of object and the location of the object in the video.
    \item The DNN developer creates a planning system which integrates the output of the DL solution with other components in the system. The planning system decides which actions the car will take.
    \item The owner acquires the planning system and produces a car in which the planning system is operational. 
    \item The end user purchases the car and uses the car to travel from point A to point B.
    \item The data subjects are all the entities from which information is captured along the route from point A to point B: pedestrians, private property such as houses, other cars. 
    \item The stakeholders are governmental institutions which formulate laws regulating the use of autonomous vehicles, or insurance companies that have to assess risk levels and their consequences.
\end{enumerate}

\section{Laws and regulations}
An important initiative within the European Union is the General Data Protection Regulation (GDPR) \footnote{\url{https://www.eugdpr.org}} that was approved on April 14, 2016, and became enforceable on May 25, 2018. The GDPR distinguishes between personal data, data subjects, data processors and data controllers (Article 4, Definitions, Paragraphs 1, 7 \& 8). \textit{Personal data} is defined as ``any information relating to an identified or identifiable natural person (data subject)''. A \textit{data processor} is the natural or legal person, public authority, agency or other body which processes data on behalf of the \textit{data controller}, who determines the purposes, conditions and means of the processing. Hence, the DNN can function as a tool to be used by the data processor, whereas owners or end users can fill the role of data controllers.

The GDPR focuses in part on profiling: ``any form of automated processing of personal data consisting of the use of personal data to evaluate certain personal aspects relating to a natural person, in particular to analyse or predict aspects concerning that natural person's performance at work, economic situation, health, personal preferences, interests, reliability, behaviour, location or movements'' (Article 4, Definitions, Paragraph 4). According to articles 13, 14 and 15, when personal data is collected from a data subject for automated decision-making, the data subject has the right to access, and the data controller is obliged to provide, ``meaningful information about the logic involved.'' Article 12 stipulates that the provision of information to data subjects should be in ``concise, transparent, intelligible and easily accessible form, using clear and plain language.''

\section{Explanation}
The right to meaningful information translates into the demand that actionable outcomes of DNNs need to be explained, i.e. be made transparent, interpretable or comprehensible to humans. Transparency refers to the extent to which an explanation makes a specific outcome understandable to a particular (group of) users. Understanding, in this context, amounts to a person grasping how a particular outcome was reached by the DNN. Note that this need not imply agreeing with the conclusion, i.e. accepting the outcome as valid or justified. In general, transparency may be considered as recommendable, leading to e.g. a greater (societal) sense of control and acceptance of ML applications. Transparency is normally also a precondition for accountability: i.e. the extent to which the responsibility for the actionable outcome can be attributed to legally (or morally) relevant agents (governments, companies, expert or lay users, etc.). However, transparency may also have negative consequences, e.g. regarding privacy or by creating possibilities for manipulation (of data, processing or training).

In relation to the (perceived) need for explanation, two reasons for investigation stand out in particular. First, a DNN may appear to dysfunction, i.e. fail to operate as intended, e.g. through bugs in the code (process malfunction). Second, it may misfunction, e.g. by producing unintended or undesired (side-)effects~\citep{floridi2015malfunctioning, mittelstadt2016ethics} that are deemed to be societally or ethically unacceptable (outcome malfunction). 
Related to dysfunction is a first category of explanations. This category is based on the information necessary in order to understand the system's basic processes, e.g. to assess whether it is functioning properly, as intended, or whether it dysfunctions (e.g. suboptimal or erroneous results). This type of explanation is normally required by DNN developers and expert users. The information is used to interpret, predict, monitor, diagnose, improve, debug or repair the functioning of a system~\citep{weller2017challenges}.

Once an application is made available to non-expert users, normally certain guarantees regarding the system’s proper functioning are in place. Generally speaking, owners, end users, data subjects and stakeholders are more interested in a second category of explanations, where suspicions about a DNN's misfunctioning (undesired outcomes) leads to requests for ``local explanations''. Users may request information about how a particular outcome was reached by the DNN, which aspects of input data, which learning factors or other parameters of the system influenced its decision or prediction. This information is then used to assess the appropriateness of the outcome in relation to the concerns and values of users~\citep{doran2017does, wachtereaan6080, doshi2017accountability, weller2017challenges}. The aim of local explanations is to strengthen the confidence and trust of users that the system is not (or will not be) conflicting with their values, i.e. that it does not violate fairness or neutrality. Note that this implies that the offered explanations should match (within certain limits) the particular user's capacity for understanding~\citep{doshi2017towards}, as indicated by the GDPR.  

\section{Explanation methods}
So far the users, the GDPR, and the role of explanations have been discussed. To bridge the gap from that area to the more technical area of explanation methods, we need to be able to evaluate the capabilities of existing methods, in the context of the users and their needs. We bridge the gap in two ways. First, we identify, on a high level, desirable properties of explanation methods. 
Second, we introduce a taxonomy to categorize all types of explanation methods and third, assess the presence of the desirable properties in the categories in our taxonomy. 

\subsection{Desirable properties of explainers}

Based on a survey of the literature, we arrive at the following properties which any {\em explainer} should have:
\begin{enumerate}
    \item \textbf{High Fidelity} The degree to which the interpretation method agrees with the input-output mapping of the DNN. This term appears in~\citep{arbatli1997rule, markowska2004rule, zilke2016deepred, ribeiro2016nothing, ribeiro2016should, andrews1995survey, lakkaraju2017interpretable}. Fidelity is arguably the most important property that an explanation model should possess. If an explanation method is not faithful to the original model then it cannot give valid explanations because the input-output mapping is incorrect. 
    
    \item \textbf{High Interpretabiliy} To what extent a user is able to obtain true insight into how actionable outcomes are obtained. We distinguish interpretability into the following two subproperties:
    \begin{enumerate}
    \item \textbf{High Clarity} The degree to which the resulting explanation is unambiguous. This property is extremely important in safety-critical applications~\citep{andrews1995survey} where ambiguity is to be avoided. \citep{lakkaraju2017interpretable} introduces a quantifiable measure of clarity (unambiguity) for their method.
    
    \item \textbf{High Parsimony} This refers to the complexity of the resulting explanation. An explanation that is parsimonious is a simple explanation. This concept is generally related to Occam's razor and in the case of explaining DNNs the principle is also of importance. The optimal degree of parsimony can in part be dependent on the user's capabilities.
    \end{enumerate}
    \item \textbf{High Generalizability} The range of architectures to which the explanation method can be applied. This increases the usefulness of the explanation method. Methods that are model-agnostic~\citep{ribeiro2016should} are the highest in generalizability.
    
    \item \textbf{High Explanatory Power} In this context this means how many phenomena the method can explain. This roughly translates to how many different kinds of questions the method can answer. Previously in Section \ref{sec:users} we have identified a number of questions that users may have. It is also linked to the notion that the explainer should be able to take a global perspective~\citep{ribeiro2016should}, in the sense that it can explain the behaviour of the model rather than only accounting for individual predictions.
\end{enumerate}

\subsection{A taxonomy for explanation methods}
\label{sec:taxonomy}
Over a relatively short period of time a plethora of explanation methods and strategies have come into existence, driven by the need of expert users to analyze and debug their DNNs. However, apart from a non-exhaustive overview of existing methods~\citep{montavon2017methods} and classification schemes for purely visual methods~\citep{grun2016taxonomy, seifert2017visualizations, zeng2016towards, kindermans2017reliability}, little is known about efforts to rigorously map the landscape of explanation methods and isolate the underlying patterns that guide explanation methods. In this section a taxonomy for explanation methods is proposed. Three main classes of explanation methods are identified and their features described. The taxonomy was derived by analyzing the historical and contemporary trends surrounding the topic of interpretation of DNNs and explainable AI. We realize that we cannot foresee the future developments of DNNs and their explainability methods. As such it is possible that in the future the taxonomy needs to be modified. We propose the following taxonomy:
\begin{description}
    \item [Rule-extraction methods] \hfill \\ 
    Extract rules that approximate the decision-making process in a DNN by utilizing the input and output of the DNN.

    \item [Attribution methods] \hfill \\ Measures the importance of a component by changing to the input or internal components and recording how much the changes affect model performance. Methods known by other names that fall in this category are occlusion, perturbation, erasure, ablation and influence. Attribution methods are often visualized and sometimes referred to as visualization methods.

    \item [Intrinsic methods] \hfill \\ Aim to improve the interpretability of internal representations with methods that are part of the DNN architecture. Intrinsic methods increase fidelity, clarity and parsimony in attribution methods.

\end{description}
In the following subsections we will describe the main features of each class and give examples from current research. 

\subsubsection{Rule-extraction methods}
Rule-extraction methods extract human interpretable rules that approximate the decision-making process in a DNN. Older genetic algorithm based rule extraction methods for ANNs can be found in~\citep{andrews1995survey, arbatli1997rule, lu2006explanatory}. \citet{andrews1995survey} specify three categories of rule extraction methods:
\begin{description}
    \item [Decompositional approach] \hfill \\ 
    Decomposition refers to breaking down the network into smaller individual parts. For the decompositional approach, the architecture of the network and/or its outputs are used in the process. 
    \citet{zilke2016deepred} uses a decompositional algorithm that extracts rules for each layer in the DNN. These rules are merged together in a final merging step to produce a set of rules that describe the network behaviour by means of its inputs. \citet{murdoch2017automatic} succeeded in extracting rules from an LSTM by applying a decompositional approach.
    \item [Pedagogical approach] \hfill \\ 
    Introduced by~\citet{craven1994using} and named by~\citet{andrews1995survey} the pedagogical approach involves ``viewing rule extraction as a learning task where the target concept is the function computed by the network and the input features are simply the network's input features''~\citep{craven1994using}. The pedagogical approach has the advantage that it is inherently model-agnostic. Recent examples are found in~\citep{ribeiro2016nothing, lakkaraju2017interpretable}. 
    \item [Eclectic approach] \hfill \\ According to~\citet{andrews1995survey} ``membership in this category is assigned to techniques which utilize knowledge about the internal architecture and/or weight vectors in the trained artificial neural network to complement a symbolic learning algorithm.''
    
\end{description}

In terms of fidelity, local explanations are more faithful than global explanations. For rule-extraction this means that rules that govern the result of a specific input, or a neighborhood of inputs are more faithful than rules that govern all possible inputs. Rule extraction is arguably the most interpretable category of methods in our taxonomy considering that the resulting set of rules can be unambiguously be interpreted by a human being as a kind of formal language. Therefore we can say that it has a high degree of clarity. In terms of parsimony we can say that if the ruleset is "small enough" the parsimony is higher than when the ruleset is ``too large". What determines ``small enough" and ``too large" is difficult to quantify formally and is also dependent on the user (expert vs. lay). In terms of generalizability it can go both ways: if a decompositional approach is used it is likely that the method is not generalizable, while if a pedagogical approach is used the method is highly generalizable. In terms of explanatory power, rule-extraction methods can 1) validate whether the network is working as expected in terms of overall logic flow, and 2) explain which aspects of the input data had an effect that lead to the specific output.

\subsubsection{Attribution methods}
Attribution, a term introduced by~\citet{ancona2018towards}, also referred to as relevance~\citep{bach2015pixel, binder2016layer, zintgraf2017visualizing, robnik2008explaining}, contribution \citep{shrikumar2017learning}, class saliency~\citep{simonyan2013deep} or influence~\citep{kindermans2016investigating, adler2016auditing, koh2017understanding}, aims to reveal components of high importance in the input to the DNN and their effect as the input is propagated through the network. Because of this property we can categorize the following methods to the attribution category: occlusion~\citep{gucluturk2017visualizing}, erasure~\citep{li2016understanding}, perturbation~\citep{fong2017interpretable}, adversarial examples~\citep{papernot2016practical} and 
prediction difference analysis \citep{zintgraf2017visualizing}. Other methods that belong to this category are found in~\citep{baehrens2010explain, murdoch2018beyond, ribeiro2016should}. It is worth mentioning that attribution methods do not only apply to image input but also to other forms of input, such as text processing by LSTMs~\citep{murdoch2018beyond}. The definition of attribution methods in this chapter is similar to that of saliency methods~\citep{kindermans2017reliability}, but more general than the definition of attribution methods in~\citep{kindermans2017reliability} akin to the definition in~\citep{ancona2018towards}. 

The majority of explanation methods for DNNs visualize the information obtained by attribution methods.
Visualization methods were popularized by~\citep{erhan2009visualizing, simonyan2013deep, zeiler2014visualizing} in recent years and are concerned with how the important features are visualized. \citet{zeng2016towards} identifies that current methods focus on three aspects of visualization: feature visualization, relationship visualization and process visualization. Overall visualization methods are very intuitive methods to gain a variety of insight about a DNN decision process on many levels including architecture assessment, model quality assessment and even user feedback integration, e.g. \citet{olah2018the} create intuitive visualization interfaces for image processing DNNs.

\citet{kindermans2017reliability} has shown recently that attribution methods ``lack reliability when the explanation is sensitive to factors that do not contribute to the model prediction.'' Furthermore they introduce the notion of \textit{input invariance} as a prerequisite for accurate attribution. In other words, if the attribution method does not satisfy input invariance, we can consider it to have low fidelity. In terms of clarity, there is a degree of ambiguity that is inherent with these methods because visual explanations can be interpreted in multiple ways by different users, even by users in the same user category. In contrast to the precise results of rule-extraction methods, the information that results from attribution methods has less structure. In addition, the degree of clarity is dependent on the degree of fidelity of the method: low fidelity can cause incorrect attribution, resulting in noisy output with distracting attributions that increase ambiguity. The degree of parsimony depends on the method of visualization itself. Methods that visualize only the significant attributions exhibit a higher degree of parsimony. The degree of generalizability depends on which components are used to determine attribution. Methods that only use the input and output are inherently model agnostic, resulting in the highest degree of generalizability. Following this logic, methods that make use of internal components are generalizable to the degree that other models share these components. For example, deconvolutional networks~\citep{zeiler2010deconvolutional} can be applied to models that make use of convolutions to extract features from input images. In terms of explanatory power, this class of methods can reflect intuitively with visual explanations which factors in the input dimension had a significant impact on the output of the DNN. However these methods do not explain the reason for the importance of the particular factor attribution. 

\subsubsection{Intrinsic methods} 
The previous categories are designed to make explainable some aspects of a DNN in a process separate from training the DNN. In contrast, this category aims to improve the interpretability of internal representations with methods that are part of the DNN architecture, e.g. as part of the loss function~\citep{dong2017improving, dong2017towards}, modules that add additional capabilities~\citep{santoro2017simple, palm2017recurrent}, or as part of the architecture structure, in terms of operations between layers~\citep{li2017aognets, wu2017interpretable, louizos2017causal, goudet2017learning}. \citet{dong2017improving} provide an interpretive loss function to increase the visual fidelity of the learned features. More importantly~\citet{dong2017towards} show that by training DNNs with adversarial data and a consistent loss, we can trace back errors made by the DNN to individual neurons and identify whether the data was adversarial. \citet{santoro2017simple} give a DNN the ability to answer relational reasoning questions about a specific environment, by introducing a relational reasoning module that learns a relational function, which can be applied to any DNN. \citet{palm2017recurrent} build on work by~\citet{santoro2017simple} and introduces a recurrent relational network which can take the temporal component into account. \citet{li2017aognets} introduce an explicit structure to DNNs for visual recognition by building in an AND-OR grammar directly in the network structure. This leads to better interpretation of the information flow in the network, hence increased parsimony in attribution methods. \citet{louizos2017causal} make use of generative neural networks perform causal inference and \citet{goudet2017learning} use generative neural networks to learn functional causal models. Intrinsic methods do not explicitly explain anything by themselves. Instead they increase fidelity, clarity and parsimony in attribution methods. This class of methods is different from attribution methods because it tries to make the DNN inherently more interpretable by changing the architecture of the DNN, where attribution methods use what is there already and only transform aspects of the representation to something meaningful after the network is trained. 

\section{Addressing general concerns}
As indicated in Figure \ref{fig:xai_components}, users have certain values, that in relation to a particular technology may lead to concerns, that in relation to particular applications can lead to specific questions. \citet{mittelstadt2016ethics} and \citet{danks2017algorithmic} distinguish various concerns that users may have. The kinds of concerns they discuss focus to a large extent on the inconclusiveness, inscrutability or misguidedness of used evidence. That is, they concern to a significant extent the reliability and accessibility of used data (data mining, generally speaking). 
In addition to apprehensions about data, there are concerns that involve aspects of the processing itself, e.g. the inferential validity of an algorithm. Also, questions may be raised about the validity of a training process (e.g. requiring information about how exactly a DNN is trained). In the following, we provide a list of general concerns that should be addressed when developing predictive models such as DNNs:

\begin{description}

 \item [\textbf{Flawed data collection}] \hfill \\
    Data collection may be flawed in several ways. Large labeled datasets that are used to train DNNs are either acquired by researchers (often via crowdsourcing) or by companies that `own' the data. However, data quality may depend on multiple factors such as noise or censoring and there is no strict control on whether data is annotated correctly. Furthermore, the characteristics of the workers who annotated the data may introduce unwanted biases~\citep{barocas2016big}. These biases may be due to preferences that do not generalize across cultures or due to stereotyping, where sensitivity to irrelevant attributes such as race or gender may induce unfair actionable outcomes. The same holds for the selection of the data that is used for annotation in the first place. Used data may reflect the status quo, which is not necessarily devoid of biases~\citep{caliskan2017semantics}. Furthermore, selection bias may have as a result that data collected in one setting need not generalize to other settings. For example, video data used to train autonomous driving systems may not generalize to other locations or conditions. 
    \item [\textbf{Inscrutable data use}] 
    \hfill \\ 
    The exact use of the data to train DNNs may also be opaque. Users may worry about what (part of the) data exactly has led to the outcome. Often it is not even known to the data subject which personal data is being used for what purposes. A case in point is the use of person data for risk profiling by governmental institutions. For example, criticisms have been raised about the way the Dutch SyRI system uses data to detect fraud.\footnote{https://bijvoorbaatverdacht.nl} Furthermore, the involvement of expert users who may be prone to biases as well may have an implicit influence on DNN training. 

    \item [\textbf{Suboptimal inferences}] 
    \hfill \\
    The inferences made by DNNs are of a correlational rather than a causal nature. This implies that subtle correlations between input features may influence network output, which themselves may be driven by various biases. Work is in progress to mitigate or remove the influence of sensitive variables that should not affect decision outcomes by embracing causal inference procedures~\citep{chiappa2018path}. Note further that the impact of suboptimal inferences is domain dependent. For example, in medicine and the social sciences, suboptimal inferences may directly affect the lives of individuals or whole populations whereas in the exact sciences, suboptimal inferences may affect evidence for or against a specific scientific theory.

    \item [\textbf{Undesirable outcomes}]
    \hfill \\
    End users or data subjects may feel that the outcome of the DNN is somehow undesirable in relation to the particular values they hold, e.g. violating fairness or privacy. Importantly, actionable outcomes should take into account preferences of the stakeholder, which can be an individual (e.g. when deciding on further medical investigation) as well as the community as a whole (e.g. in case of policies about autonomous driving or predictive policing). These considerations demand the involvement of domain experts and ethicists already in the earliest stages of model development. Finally, model predictions may be of a statistical rather than deterministic nature. This speaks for the inclusion of decision-theoretic constructs in deciding on optimal actionable outcomes~\citep{VonNeumann1953}.

    \item [\textbf{Adversarial attacks}]
    \hfill \\
    Images~\citep{szegedy2013intriguing, cubuk2017intriguing} and audio~\citep{carlini2018audio} can easily be distorted with modifications that are imperceptible to humans. Such distortions cause DNNs to make incorrect inferences and can be done with the purpose of intentionally misleading DNNs (e.g. yielding predictions in favor of the perpetrator). Work in progress shows that there are methods to detect adversarial instances~\citep{rawat2017adversarial} and to mitigate the attacks~\citep{lin2017detecting}. However further research is needed to increase the robustness of DNNs against adversarial attacks as there are no methods in existence that fully diminish the effects of adversarial attacks.

\end{description}

As stated by~\citet{doran2017does}, explanation methods may make predictive models such as DNNs more comprehensible. However, explanation methods alone not completely resolve the raised concerns.

\section{Discussion}
\label{sec:discussion}
In this chapter we set out to analyze the question of ``What can be explained?'' given the users and their needs, laws and regulations, and existing explanation methods. Specifically, we looked at the capabilities of explanation methods and analyzed which questions/concerns about explainability these methods address in the context of DNNs. 

Overall, it is clear that (visual) explanations can be given about various aspects of the influence of the input on the output (e.g. given the input data, which aspects of the data lead to the output?), by making use of both  rule-extraction and attribution methods. Also, when used in combination with attribution methods, intrinsic methods lead to more explainable DNNs. It is likely that in the future we will see the rise of a new category of explanation methods that combine aspects of rule-extraction, attribution and intrinsic methods, to answer specific questions in a simple human interpretable language. 

Furthermore, it is obvious that current explanation methods are tailored to expert users, since the interpretation of the results require knowledge of the DNN process. As far as we are aware, explanation methods, e.g. intuitive explanation interfaces, for lay users do not exist. Ideally, if such explanation methods would exist, they should be able to answer, in a simple human language, questions about every operation that the application performs. This is not an easy task since the number of conceivable questions one could ask about the working of an application is substantial. 

Two particular concerns, which are difficult to address with explanation methods, is the concern about bias in datasets that leads to biased DNNs, as well as the suspicion about unfair outcomes: Can we indicate that the DNN is biased, and if so, can we remove the bias? Has the DNN been applied responsibly? These are not problems that are directly solvable with explanation methods. However, explanation methods alleviate the first problem to the extent that learned features can be visualized (using attribution methods) and further analyzed for bias using other methods that are not explanation methods. For the second problem, more general measures, such as regulations and laws, will need to be developed.


\begin{thebibliography}{99}
\bibitem[Adler et~al.(2016)Adler, Falk, Friedler, Rybeck, Scheidegger, Smith,
  and Venkatasubramanian]{adler2016auditing}
Philip Adler, Casey Falk, Sorelle~A. Friedler, Gabriel Rybeck, Carlos
  Scheidegger, Brandon Smith, and Suresh Venkatasubramanian.
\newblock Auditing black-box models for indirect influence.
\newblock In \emph{2016 {IEEE} 16th International Conference on Data Mining
  ({ICDM})}. {IEEE}, 2016.

\bibitem[Ancona et~al.(2018)Ancona, Ceolini, Oztireli, and
  Gross]{ancona2018towards}
Marco Ancona, Enea Ceolini, Cengiz Oztireli, and Markus Gross.
\newblock Towards better understanding of gradient-based attribution methods
  for deep neural networks.
\newblock In \emph{6th International Conference on Learning Representations
  (ICLR 2018)}, 2018.

\bibitem[Andrews et~al.(1995)Andrews, Diederich, and Tickle]{andrews1995survey}
Robert Andrews, Joachim Diederich, and Alan~B. Tickle.
\newblock Survey and critique of techniques for extracting rules from trained
  artificial neural networks.
\newblock \emph{Knowledge-Based Systems}, 8\penalty0 (6):\penalty0 373--389,
  1995.

\bibitem[Arbatli and Akin(1997)]{arbatli1997rule}
A.~Duygu Arbatli and H.~Levent Akin.
\newblock Rule extraction from trained neural networks using genetic
  algorithms.
\newblock \emph{Nonlinear Analysis: Theory, Methods {\&} Applications},
  30\penalty0 (3):\penalty0 1639--1648, 1997.

\bibitem[Bach et~al.(2015)Bach, Binder, Montavon, Klauschen, M{\"u}ller, and
  Samek]{bach2015pixel}
Sebastian Bach, Alexander Binder, Gr{\'e}goire Montavon, Frederick Klauschen,
  Klaus-Robert M{\"u}ller, and Wojciech Samek.
\newblock On pixel-wise explanations for non-linear classifier decisions by
  layer-wise relevance propagation.
\newblock \emph{{PLOS} {ONE}}, 10\penalty0 (7), 2015.

\bibitem[Baehrens et~al.(2010)Baehrens, Schroeter, Harmeling, Kawanabe, Hansen,
  and M{\"u}ller]{baehrens2010explain}
David Baehrens, Timon Schroeter, Stefan Harmeling, Motoaki Kawanabe, Katja
  Hansen, and Klaus-Robert M{\"u}ller.
\newblock How to explain individual classification decisions.
\newblock \emph{Journal of Machine Learning Research ({JMLR})}, 11:\penalty0
  1803--1831, 2010.

\bibitem[Barocas and Selbst(2016)]{barocas2016big}
Solon Barocas and Andrew~D. Selbst.
\newblock Big data's disparate impact.
\newblock \emph{Cal. L. Rev.}, 104:\penalty0 671, 2016.

\bibitem[Binder et~al.(2016)Binder, Bach, Montavon, M{\"u}ller, and
  Samek]{binder2016layer}
Alexander Binder, Sebastian Bach, Gr{\'e}goire Montavon, Klaus-Robert
  M{\"u}ller, and Wojciech Samek.
\newblock Layer-wise relevance propagation for deep neural network
  architectures.
\newblock In \emph{Information Science and Applications (ICISA) 2016}, pages
  913--922. Springer, 2016.

\bibitem[Bojarski et~al.(2017)Bojarski, Yeres, Choromanska, Choromanski,
  Firner, Jackel, and Muller]{bojarski2017explaining}
Mariusz Bojarski, Philip Yeres, Anna Choromanska, Krzysztof Choromanski,
  Bernhard Firner, Lawrence Jackel, and Urs Muller.
\newblock Explaining how a deep neural network trained with end-to-end learning
  steers a car.
\newblock \emph{arXiv preprint arXiv:1704.07911}, 2017.

\bibitem[Caliskan et~al.(2017)Caliskan, Bryson, and
  Narayanan]{caliskan2017semantics}
Aylin Caliskan, Joanna~J. Bryson, and Arvind Narayanan.
\newblock Semantics derived automatically from language corpora contain
  human-like biases.
\newblock \emph{Science}, 356\penalty0 (6334):\penalty0 183--186, 2017.

\bibitem[Carlini and Wagner(2018)]{carlini2018audio}
Nicholas Carlini and David Wagner.
\newblock Audio adversarial examples: Targeted attacks on speech-to-text.
\newblock \emph{arXiv preprint arXiv:1801.01944}, 2018.

\bibitem[Chiappa and Gillam(2018)]{chiappa2018path}
Silvia Chiappa and Thomas~P.S. Gillam.
\newblock Path-specific counterfactual fairness.
\newblock \emph{arXiv preprint arXiv:1802.08139}, 2018.

\bibitem[Craven and Shavlik(1994)]{craven1994using}
Mark~W. Craven and Jude~W. Shavlik.
\newblock Using sampling and queries to extract rules from trained neural
  networks.
\newblock In \emph{Machine Learning Proceedings 1994}, pages 37--45. Elsevier,
  1994.

\bibitem[Cubuk et~al.(2017)Cubuk, Zoph, Schoenholz, and
  Le]{cubuk2017intriguing}
Ekin~D. Cubuk, Barret Zoph, Samuel~S. Schoenholz, and Quoc~V. Le.
\newblock Intriguing properties of adversarial examples.
\newblock \emph{arXiv preprint arXiv:1711.02846}, 2017.

\bibitem[Danks and London(2017)]{danks2017algorithmic}
David Danks and Alex~John London.
\newblock Algorithmic bias in autonomous systems.
\newblock In \emph{Proceedings of the 26th International Joint Conference on
  Artificial Intelligence ({IJCAI})}, pages 4691--4697. AAAI Press, 2017.

\bibitem[Dong et~al.(2017{\natexlab{a}})Dong, Su, Zhu, and
  Bao]{dong2017towards}
Yinpeng Dong, Hang Su, Jun Zhu, and Fan Bao.
\newblock Towards interpretable deep neural networks by leveraging adversarial
  examples.
\newblock \emph{arXiv preprint arXiv:1708.05493}, 2017{\natexlab{a}}.

\bibitem[Dong et~al.(2017{\natexlab{b}})Dong, Su, Zhu, and
  Zhang]{dong2017improving}
Yinpeng Dong, Hang Su, Jun Zhu, and Bo~Zhang.
\newblock Improving interpretability of deep neural networks with semantic
  information.
\newblock In \emph{2017 {IEEE} Conference on Computer Vision and Pattern
  Recognition ({CVPR})}. {IEEE}, 2017{\natexlab{b}}.

\bibitem[Doran et~al.(2017)Doran, Schulz, and Besold]{doran2017does}
Derek Doran, Sarah Schulz, and Tarek~R. Besold.
\newblock What does explainable {AI} really mean? a new conceptualization of
  perspectives.
\newblock \emph{arXiv preprint arXiv:1710.00794}, 2017.

\bibitem[Doshi-Velez and Kim(2017)]{doshi2017towards}
Finale Doshi-Velez and Been Kim.
\newblock Towards a rigorous science of interpretable machine learning.
\newblock \emph{arXiv preprint arXiv:1702.08608}, 2017.

\bibitem[Doshi-Velez et~al.(2017)Doshi-Velez, Kortz, Budish, Bavitz, Gershman,
  O{'}Brien, Shieber, Waldo, Weinberger, and
  Wood]{doshi2017accountability}
Finale Doshi-Velez, Mason Kortz, Ryan Budish, Christopher Bavitz, Samuel~J.
  Gershman, David O{'}Brien, Stuart Shieber, Jim Waldo, David
  Weinberger, and Alexandra Wood.
\newblock Accountability of {AI} under the law: The role of explanation.
\newblock \emph{{SSRN} Electronic Journal}, 2017.

\bibitem[Elman(1989)]{elman1989representation}
Jeffrey~L. Elman.
\newblock Representation and structure in connectionist models.
\newblock Technical report, 1989.

\bibitem[Erhan et~al.(2009)Erhan, Bengio, Courville, and
  Vincent]{erhan2009visualizing}
Dumitru Erhan, Yoshua Bengio, Aaron Courville, and Pascal Vincent.
\newblock Visualizing higher-layer features of a deep network.
\newblock \emph{University of Montreal}, 1341:\penalty0 3, 2009.

\bibitem[Floridi et~al.(2015)Floridi, Fresco, and
  Primiero]{floridi2015malfunctioning}
Luciano Floridi, Nir Fresco, and Giuseppe Primiero.
\newblock On malfunctioning software.
\newblock \emph{Synthese}, 192\penalty0 (4):\penalty0 1199--1220, 2015.

\bibitem[Fong and Vedaldi(2017)]{fong2017interpretable}
Ruth~C. Fong and Andrea Vedaldi.
\newblock Interpretable explanations of black boxes by meaningful perturbation.
\newblock In \emph{2017 {IEEE} International Conference on Computer Vision
  ({ICCV})}. {IEEE}, 2017.

\bibitem[Goudet et~al.(2017)Goudet, Kalainathan, Caillou, Lopez-Paz, Guyon,
  Sebag, Tritas, and Tubaro]{goudet2017learning}
Olivier Goudet, Diviyan Kalainathan, Philippe Caillou, David Lopez-Paz,
  Isabelle Guyon, Michele Sebag, Aris Tritas, and Paola Tubaro.
\newblock Learning functional causal models with generative neural networks.
\newblock \emph{arXiv preprint arXiv:1709.05321}, 2017.

\bibitem[Gr{\"u}n et~al.(2016)Gr{\"u}n, Rupprecht, Navab, and
  Tombari]{grun2016taxonomy}
Felix Gr{\"u}n, Christian Rupprecht, Nassir Navab, and Federico Tombari.
\newblock A taxonomy and library for visualizing learned features in
  convolutional neural networks.
\newblock \emph{arXiv preprint arXiv:1606.07757}, 2016.

\bibitem[Gu\c{c}l\"{u}t\"{u}rk et~al.(2017)Gu\c{c}l\"{u}t\"{u}rk,
  G\"{u}\c{c}l\"{u}, Perez, Jair~Escalante, Baro, Guyon, Andujar,
  Jacques~Junior, Madadi, Escalera, van Gerven, and van
  Lier]{gucluturk2017visualizing}
Ya\v{g}mur Gu\c{c}l\"{u}t\"{u}rk, Umut G\"{u}\c{c}l\"{u}, Marc Perez, Hugo
  Jair~Escalante, Xavier Baro, Isabelle Guyon, Carlos Andujar, Julio
  Jacques~Junior, Meysam Madadi, Sergio Escalera, Marcel A.~J. van Gerven, and
  Rob van Lier.
\newblock Visualizing apparent personality analysis with deep residual
  networks.
\newblock In \emph{2017 {IEEE} International Conference on Computer Vision
  Workshops ({ICCVW})}, pages 3101--3109, 2017.

\bibitem[Gunning(2017)]{gunning2017explainable}
David Gunning.
\newblock Explainable artificial intelligence ({XAI}).
\newblock \emph{Defense Advanced Research Projects Agency (DARPA)}, 2017.

\bibitem[Hall et~al.(2017)Hall, Phan, and Ambati]{hall2017ideas}
Patrick Hall, Wen Phan, and SriSatish Ambati.
\newblock Ideas on interpreting machine learning.
\newblock 2017.
\newblock Available online at:
  https://www.oreilly.com/ideas/ideas-on-interpreting-machine-learning.

\bibitem[Holzinger et~al.(2017{\natexlab{a}})Holzinger, Biemann, Pattichis, and
  Kell]{holzinger2017we}
Andreas Holzinger, Chris Biemann, Constantinos~S. Pattichis, and Douglas~B.
  Kell.
\newblock What do we need to build explainable {AI} systems for the medical
  domain?
\newblock \emph{arXiv preprint arXiv:1712.09923}, 2017{\natexlab{a}}.

\bibitem[Holzinger et~al.(2017{\natexlab{b}})Holzinger, Plass, Holzinger,
  Cri{\c{s}}an, Pintea, and Palade]{holzinger2017glass}
Andreas Holzinger, Markus Plass, Katharina Holzinger, Gloria~Cerasela
  Cri{\c{s}}an, Camelia-Mihaela Pintea, and Vasile Palade.
\newblock A glass-box interactive machine learning approach for solving
  {NP}-hard problems with the human-in-the-loop.
\newblock \emph{arXiv preprint arXiv:1708.01104}, 2017{\natexlab{b}}.

\bibitem[Kindermans et~al.(2016)Kindermans, Sch{\"u}tt, M{\"u}ller, and
  D{\"a}hne]{kindermans2016investigating}
Pieter-Jan Kindermans, Kristof~T. Sch{\"u}tt, Klaus-Robert M{\"u}ller, and Sven
  D{\"a}hne.
\newblock Investigating the influence of noise and distractors on the
  interpretation of neural networks.
\newblock \emph{arXiv preprint arXiv:1611.07270}, 2016.

\bibitem[Kindermans et~al.(2017)Kindermans, Hooker, Adebayo, Alber, Sch{\"u}tt,
  D{\"a}hne, Erhan, and Kim]{kindermans2017reliability}
Pieter-Jan Kindermans, Sara Hooker, Julius Adebayo, Maximilian Alber,
  Kristof~T. Sch{\"u}tt, Sven D{\"a}hne, Dumitru Erhan, and Been Kim.
\newblock The (un) reliability of saliency methods.
\newblock \emph{arXiv preprint arXiv:1711.00867}, 2017.

\bibitem[Koh and Liang(2017)]{koh2017understanding}
Pang~Wei Koh and Percy Liang.
\newblock Understanding black-box predictions via influence functions.
\newblock In \emph{Proceedings of the 34th International Conference on Machine
  Learning ({ICML})}, volume~70 of \emph{Proceedings of Machine Learning
  Research ({PMLR})}, pages 1885--1894, 2017.

\bibitem[Lakkaraju et~al.(2017)Lakkaraju, Kamar, Caruana, and
  Leskovec]{lakkaraju2017interpretable}
Himabindu Lakkaraju, Ece Kamar, Rich Caruana, and Jure Leskovec.
\newblock Interpretable \& explorable approximations of black box models.
\newblock \emph{arXiv preprint arXiv:1707.01154}, 2017.

\bibitem[Lee et~al.(2017)Lee, Tajmir, Lee, Zissen, Yeshiwas, Alkasab, Choy, and
  Do]{lee2017fully}
Hyunkwang Lee, Shahein Tajmir, Jenny Lee, Maurice Zissen, Bethel~Ayele
  Yeshiwas, Tarik~K. Alkasab, Garry Choy, and Synho Do.
\newblock Fully automated deep learning system for bone age assessment.
\newblock \emph{Journal of Digital Imaging ({JDI})}, 30\penalty0 (4):\penalty0
  427--441, 2017.

\bibitem[Li et~al.(2016)Li, Monroe, and Jurafsky]{li2016understanding}
Jiwei Li, Will Monroe, and Dan Jurafsky.
\newblock Understanding neural networks through representation erasure.
\newblock \emph{arXiv preprint arXiv:1612.08220}, 2016.

\bibitem[Li et~al.(2017)Li, Wu, Song, and Krim]{li2017aognets}
Xilai Li, Tianfu Wu, Xi~Song, and Hamid Krim.
\newblock {AOGNets}: Deep {AND-OR} grammar networks for visual recognition.
\newblock \emph{arXiv preprint arXiv:1711.05847}, 2017.

\bibitem[Lin et~al.(2017)Lin, Liu, Sun, and Huang]{lin2017detecting}
Yen-Chen Lin, Ming-Yu Liu, Min Sun, and Jia-Bin Huang.
\newblock Detecting adversarial attacks on neural network policies with visual
  foresight.
\newblock \emph{arXiv preprint arXiv:1710.00814}, 2017.

\bibitem[Lockett et~al.()Lockett, Jefferies, Etheridge, and
  Brewer]{lockett2017predictions}
Alan Lockett, Trevor Jefferies, Neil Etheridge, and Alicia Brewer.
\newblock White paper tag predictions: How {DISCO AI} is bringing deep learning
  to legal technology.
\newblock Available online at: https://www.csdisco.com/disco-ai.

\bibitem[Louizos et~al.(2017)Louizos, Shalit, Mooij, Sontag, Zemel, and
  Welling]{louizos2017causal}
Christos Louizos, Uri Shalit, Joris~M. Mooij, David Sontag, Richard Zemel, and
  Max Welling.
\newblock Causal effect inference with deep latent-variable models.
\newblock In \emph{Advances in Neural Information Processing Systems 30
  ({NIPS})}, pages 6446--6456. 2017.

\bibitem[Lu et~al.(2006)Lu, Tokinaga, and Ikeda]{lu2006explanatory}
Jianjun Lu, Shozo Tokinaga, and Yoshikazu Ikeda.
\newblock Explanatory rule extraction based on the trained neural network and
  the genetic programming.
\newblock \emph{Journal of the Operations Research Society of Japan ({JORSJ})},
  49\penalty0 (1):\penalty0 66--82, 2006.

\bibitem[Marcus(2018)]{marcus2018deep}
Gary Marcus.
\newblock Deep learning: A critical appraisal.
\newblock \emph{arXiv preprint arXiv:1801.00631}, 2018.

\bibitem[Markowska-Kaczmar and Wnuk-Lipi{\'n}ski(2004)]{markowska2004rule}
Urszula Markowska-Kaczmar and Pawe{\l} Wnuk-Lipi{\'n}ski.
\newblock Rule extraction from neural network by genetic algorithm with pareto
  optimization.
\newblock \emph{Artificial Intelligence and Soft Computing-ICAISC 2004}, pages
  450--455, 2004.

\bibitem[Mittelstadt et~al.(2016)Mittelstadt, Allo, Taddeo, Wachter, and
  Floridi]{mittelstadt2016ethics}
Brent~Daniel Mittelstadt, Patrick Allo, Mariarosaria Taddeo, Sandra Wachter,
  and Luciano Floridi.
\newblock The ethics of algorithms: Mapping the debate.
\newblock \emph{Big Data \& Society}, 3\penalty0 (2), 2016.

\bibitem[Montavon et~al.(2018)Montavon, Samek, and
  M{\"u}ller]{montavon2017methods}
Gr{\'e}goire Montavon, Wojciech Samek, and Klaus-Robert M{\"u}ller.
\newblock Methods for interpreting and understanding deep neural networks.
\newblock \emph{Digital Signal Processing}, 73:\penalty0 1--15, 2018.

\bibitem[Murdoch and Szlam(2017)]{murdoch2017automatic}
W.~James Murdoch and Arthur Szlam.
\newblock Automatic rule extraction from long short term memory networks.
\newblock In \emph{International Conference on Learning Representations
  (ICLR)}, 2017.

\bibitem[Murdoch et~al.(2018)Murdoch, Liu, and Yu]{murdoch2018beyond}
W.~James Murdoch, Peter~J. Liu, and Bin Yu.
\newblock Beyond word importance: Contextual decomposition to extract
  interactions from {LSTMs}.
\newblock In \emph{International Conference on Learning Representations
  ({ICLR})}, 2018.

\bibitem[Olah et~al.(2017)Olah, Mordvintsev, and Schubert]{olah2017feature}
Chris Olah, Alexander Mordvintsev, and Ludwig Schubert.
\newblock Feature visualization.
\newblock \emph{Distill}, 2017.
\newblock Available online at: https://distill.pub/2017/feature-visualization.

\bibitem[Olah et~al.(2018)Olah, Satyanarayan, Johnson, Carter, Schubert, Ye,
  and Mordvintsev]{olah2018the}
Chris Olah, Arvind Satyanarayan, Ian Johnson, Shan Carter, Ludwig Schubert,
  Katherine Ye, and Alexander Mordvintsev.
\newblock The building blocks of interpretability.
\newblock \emph{Distill}, 2018.
\newblock Available online at: https://distill.pub/2018/building-blocks.

\bibitem[Palm et~al.(2017)Palm, Paquet, and Winther]{palm2017recurrent}
Rasmus~Berg Palm, Ulrich Paquet, and Ole Winther.
\newblock Recurrent relational networks for complex relational reasoning.
\newblock \emph{arXiv preprint arXiv:1711.08028}, 2017.

\bibitem[Papernot et~al.(2017)Papernot, McDaniel, Goodfellow, Jha, Celik, and
  Swami]{papernot2016practical}
Nicolas Papernot, Patrick McDaniel, Ian Goodfellow, Somesh Jha, Z.~Berkay
  Celik, and Ananthram Swami.
\newblock Practical black-box attacks against machine learning.
\newblock In \emph{Proceedings of the 2017 ACM on Asia Conference on Computer
  and Communications Security ({ASIA CCS '17})}, pages 506--519, 2017.

\bibitem[Rawat et~al.(2017)Rawat, Wistuba, and Nicolae]{rawat2017adversarial}
Ambrish Rawat, Martin Wistuba, and Maria-Irina Nicolae.
\newblock Adversarial phenomenon in the eyes of bayesian deep learning.
\newblock \emph{arXiv preprint arXiv:1711.08244}, 2017.

\bibitem[Ribeiro et~al.(2016{\natexlab{a}})Ribeiro, Singh, and
  Guestrin]{ribeiro2016nothing}
Marco~Tulio Ribeiro, Sameer Singh, and Carlos Guestrin.
\newblock Nothing else matters: Model-agnostic explanations by identifying
  prediction invariance.
\newblock In \emph{NIPS Workshop on Interpretable Machine Learning in Complex
  Systems}, 2016{\natexlab{a}}.

\bibitem[Ribeiro et~al.(2016{\natexlab{b}})Ribeiro, Singh, and
  Guestrin]{ribeiro2016should}
Marco~Tulio Ribeiro, Sameer Singh, and Carlos Guestrin.
\newblock "{Why} should {I} trust you?" {E}xplaining the predictions of any
  classifier.
\newblock In \emph{Proceedings of the 22nd ACM SIGKDD International Conference
  on Knowledge Discovery and Data Mining ({KDD '16})}, pages 1135--1144,
  2016{\natexlab{b}}.

\bibitem[Robnik-{\v{S}}ikonja and Kononenko(2008)]{robnik2008explaining}
Marko Robnik-{\v{S}}ikonja and Igor Kononenko.
\newblock Explaining classifications for individual instances.
\newblock \emph{IEEE Transactions on Knowledge and Data Engineering},
  20\penalty0 (5):\penalty0 589--600, 2008.

\bibitem[Samek et~al.(2017)Samek, Wiegand, and
  M{\"u}ller]{samek2017explainable}
Wojciech Samek, Thomas Wiegand, and Klaus-Robert M{\"u}ller.
\newblock Explainable artificial intelligence: Understanding, visualizing and
  interpreting deep learning models.
\newblock \emph{arXiv preprint arXiv:1708.08296}, 2017.

\bibitem[Santoro et~al.(2017)Santoro, Raposo, Barrett, Malinowski, Pascanu,
  Battaglia, and Lillicrap]{santoro2017simple}
Adam Santoro, David Raposo, David~G.T. Barrett, Mateusz Malinowski, Razvan
  Pascanu, Peter Battaglia, and Timothy Lillicrap.
\newblock A simple neural network module for relational reasoning.
\newblock \emph{arXiv preprint arXiv:1706.01427}, 2017.

\bibitem[Seifert et~al.(2017)Seifert, Aamir, Balagopalan, Jain, Sharma,
  Grottel, and Gumhold]{seifert2017visualizations}
Christin Seifert, Aisha Aamir, Aparna Balagopalan, Dhruv Jain, Abhinav Sharma,
  Sebastian Grottel, and Stefan Gumhold.
\newblock Visualizations of deep neural networks in computer vision: A survey.
\newblock In \emph{Transparent Data Mining for Big and Small Data}, pages
  123--144. Springer, 2017.

\bibitem[Shrikumar et~al.(2017)Shrikumar, Greenside, and
  Kundaje]{shrikumar2017learning}
Avanti Shrikumar, Peyton Greenside, and Anshul Kundaje.
\newblock Learning important features through propagating activation
  differences.
\newblock In \emph{Proceedings of the 34th International Conference on Machine
  Learning ({ICML})}, volume~70 of \emph{Proceedings of Machine Learning
  Research ({PMLR})}, 2017.

\bibitem[Simonyan et~al.(2013)Simonyan, Vedaldi, and
  Zisserman]{simonyan2013deep}
Karen Simonyan, Andrea Vedaldi, and Andrew Zisserman.
\newblock Deep inside convolutional networks: Visualising image classification
  models and saliency maps.
\newblock \emph{arXiv preprint arXiv:1312.6034}, 2013.

\bibitem[Szegedy et~al.(2013)Szegedy, Zaremba, Sutskever, Bruna, Erhan,
  Goodfellow, and Fergus]{szegedy2013intriguing}
Christian Szegedy, Wojciech Zaremba, Ilya Sutskever, Joan Bruna, Dumitru Erhan,
  Ian Goodfellow, and Rob Fergus.
\newblock Intriguing properties of neural networks.
\newblock \emph{arXiv preprint arXiv:1312.6199}, 2013.

\bibitem[von Neumann and Morgenstern(1953)]{VonNeumann1953}
John von Neumann and Oskar Morgenstern.
\newblock \emph{{Theory of Games and Economic Behavior}}.
\newblock Princeton University Press, Princeton, NJ, 3rd edition, 1953.

\bibitem[Wachter et~al.(2017)Wachter, Mittelstadt, and
  Floridi]{wachtereaan6080}
Sandra Wachter, Brent Mittelstadt, and Luciano Floridi.
\newblock Transparent, explainable, and accountable {AI} for robotics.
\newblock \emph{Science Robotics}, 2\penalty0 (6), 2017.

\bibitem[Weller(2017)]{weller2017challenges}
Adrian Weller.
\newblock Challenges for transparency.
\newblock \emph{Workshop on Human Interpretability in Machine Learning -- ICML
  2017}, 2017.

\bibitem[Wu et~al.(2017)Wu, Li, Song, Sun, Dong, and Li]{wu2017interpretable}
Tianfu Wu, Xilai Li, Xi~Song, Wei Sun, Liang Dong, and Bo~Li.
\newblock Interpretable {R-CNN}.
\newblock \emph{arXiv preprint arXiv:1711.05226}, 2017.

\bibitem[Zeiler and Fergus(2014)]{zeiler2014visualizing}
Matthew~D. Zeiler and Rob Fergus.
\newblock Visualizing and understanding convolutional networks.
\newblock In \emph{European Conference on Computer Vision ({ECCV})}, pages
  818--833. Springer, 2014.

\bibitem[Zeiler et~al.(2010)Zeiler, Krishnan, Taylor, and
  Fergus]{zeiler2010deconvolutional}
Matthew~D. Zeiler, Dilip Krishnan, Graham~W. Taylor, and Rob Fergus.
\newblock Deconvolutional networks.
\newblock In \emph{2010 IEEE Conference on Computer Vision and Pattern
  Recognition (CVPR)}, pages 2528--2535. IEEE, 2010.

\bibitem[Zeng(2016)]{zeng2016towards}
Haipeng Zeng.
\newblock Towards better understanding of deep learning with visualization.
\newblock 2016.

\bibitem[Zilke et~al.(2016)Zilke, Menc{\'\i}a, and Janssen]{zilke2016deepred}
Jan~Ruben Zilke, Eneldo~Loza Menc{\'\i}a, and Frederik Janssen.
\newblock Deep{RED} -- {R}ule extraction from deep neural networks.
\newblock In \emph{International Conference on Discovery Science ({ICDS})},
  pages 457--473. Springer, 2016.

\bibitem[Zintgraf et~al.(2017)Zintgraf, Cohen, Adel, and
  Welling]{zintgraf2017visualizing}
Luisa~M. Zintgraf, Taco~S. Cohen, Tameem Adel, and Max Welling.
\newblock Visualizing deep neural network decisions: Prediction difference
  analysis.
\newblock In \emph{International Conference on Learning Representations
  (ICLR)}, 2017.
  
\end{thebibliography}
\end{document}